\documentclass{article}

\usepackage[final]{corl_2017} 
\usepackage[utf8]{inputenc}
\usepackage[english]{babel}

\usepackage{url, amsmath, amssymb, booktabs, graphicx}
\graphicspath{{images/}}
\usepackage[export]{adjustbox}
\usepackage{lipsum}
\usepackage{eurosym}
\usepackage{graphicx}
\usepackage{multirow}

\usepackage{algorithm} 
\usepackage{program} 
\usepackage{algorithmic}
\usepackage{listings}
\usepackage[normalem]{ulem}

\usepackage{caption}
\DeclareCaptionFormat{listing}{\rule{\dimexpr0.9\columnwidth+17pt\relax}{0.4pt}\par\vskip1pt#1#2#3}

\makeatletter
\newenvironment{code}[1][htb]{%
    \renewcommand{\ALG@name}{Code Listing}
   \begin{algorithm}[#1]%
  }{\end{algorithm}}
\makeatother

\DeclareMathVersion{sans}
\SetSymbolFont{operators}{sans}{OT1}{cmbr}{m}{n}
\SetSymbolFont{letters}  {sans}{OML}{cmbrm}{m}{it}
\SetSymbolFont{symbols}  {sans}{OMS}{cmbrs}{m}{n}

\lstnewenvironment{sflisting}[1][]
  {\lstset{#1}\mathversion{sans}}{}

\title{robot\_gym: accelerated robot training through simulation in the cloud with ROS and Gazebo}

 \author{
   Víctor Mayoral Vilches,
   Alejandro Hernández Cordero,
   Asier Bilbao Calvo, \\
   \textbf{Irati Zamalloa Ugarte},
   \textbf{Risto Kojcev} \\
   Erle Robotics S.L. \\
   Vitoria-Gasteiz, Álava
   Spain \\
   \texttt{contact@erlerobotics.com} \\
 }

\begin{document}
\maketitle

\vspace{-1em}
\begin{abstract}

Rather than programming, training allows robots to achieve behaviors that generalize better and are capable to respond to real-world needs. However, such training requires a big amount of experimentation which is not always feasible for a physical robot. In this work, we present \emph{robot\_gym}, a framework to accelerate robot training through simulation in the cloud that makes use of roboticists' tools, simplifying the development and deployment processes on real robots. We unveil that, for simple tasks, simple 3DoF robots require more than 140 attempts to learn. For more complex, 6DoF robots, the number of attempts increases to more than 900 for the same task. We demonstrate that our framework, for simple tasks, accelerates the robot training time by more than 33\% while maintaining similar levels of accuracy and repeatability.



\end{abstract}

\keywords{Simulation, Reinforcement Learning, Policy Optimization, Robot Operating System, Gazebo, Cloud computing.} 


\section{Introduction}
\label{sec:intro}

Reinforcement Learning (RL) has recently gained attention in the robotics field. Rather than programming, it allows roboticists to train robots, producing results that generalize better and are able to comply with the dynamic environments typically encountered in robotics. Furthermore, RL techniques, if used in combination with modular robotics, could empower a new generation of robots that are more adaptable and capable of performing a variety of tasks without human intervention \cite{2018arXiv180204082M}.

While some results showed the feasibility of using RL in real robots \cite{2016arXiv160302199L}, such approach is expensive, requires hundreds of thousands of attempts (performed by a group of robots) and a period of several months. These capabilities are only available to a restricted few, thereby training in simulation has gained popularity. The idea behind using simulation is to train a virtual model of the real robot until the desired behavior is learned and then transfer the knowledge to the real robot. The behavior can be further enhanced by exposing it to a restricted number of additional training iterations. Following some the initial releases of OpenAI's gym\cite{1606.01540}, many groups started using the Mujoco\cite{todorov2012mujoco} physics engine. Others have used the Gazebo robot simulator\cite{koenig2004design} in combination with the Robot Operating System (ROS)\cite{Quigley09} to create an environment with the common tools used by roboticists named \emph{gym\_gazebo} \cite{2016arXiv160805742Z}.

In this work, we introduce an extension of \emph{gym\_gazebo}, called \emph{robot\_gym}, that makes use of container technology to deploy experiments in a distributed way, accelerating the training process through a framework for distributed RL. We aim to provide answers to the following questions: \emph{By how much is it possible to accelerate robot training time with RL techniques?} And: \emph{What is the associated cost of doing so?} Our experimental results show a significant reduction of the training time. Compared to standard RL approaches, we achieve time reductions of up to 50\% for simple tasks.


The remainder of the text is organized as follows: Section \ref{sec:previous} covers previous work. Section \ref{sec:robotgym} describes the \emph{robot\_gym} framework and section \ref{sec:experimental} the experimental results obtained when applying the framework to two different robot arms. Finally, section \ref{sec:conclusions} discusses about next steps and conclusions from this work.


\section{Previous work}
\label{sec:previous}




The work of Levine et. al \cite{gu2017deep, yahya2017collective} argues that RL, compared to the standard approach of programming robots, has potential for robots to learn a variety of skills with minimal human intervention. However, the largest bottleneck for applying RL techniques to real robot systems is the amount of training time and hand-crafted policy for the robot to learn certain behaviour. They developed an approach which parallelizes the deep Q-functions as well as the guided policy search  across several real robotic systems. Their results demonstrate that it is possible for the robot to learn a variety of complex tasks, such as grasping or door opening. This approach shows that it can achieve better generalization, utilization and training times than the single robot alternative. 

According to Liang et al. \cite{2017arXiv171209381L}, some of the most relevant problems in reinforcement learning stem from the need to scale learning and simulation while also integrating a rapidly increasing range of algorithms and models. In particular, it is stated that there is a fundamental need for composable parallel primitives to support research in RL. This statement follows from the fact that there is a growing number of RL libraries available \cite{caspi_itai_2017_1134899, pytorchrl, baselines, 2017arXiv170902878H}. Yet, very few enable the composition of components at scale.  The authors argue that these libraries offer parallelism at the level of the entire program, coupling all algorithm components together and making existing implementations difficult to scale, combine, and reuse. Based on this reasoning, the authors argue for distributing RL components in a composable way by adapting algorithms for top-down hierarchical control, thereby encapsulating parallelism and resource requirements within short-running compute tasks. They implement it on top of flexible task-based programming models like Ray\cite{2017arXiv171205889M}, which allow subroutines to be scheduled and executed asynchronously on worker processes, on a fine-grained basis, and for results to be retrieved or passed between processes.

Stooke et al. \cite{stooke2018accelerated} claim that the experiment turn-around time remains a key bottleneck in Deep Reinforcement Learning (DRL), both in research and in practice. They confirm that both, policy gradient and value iteration algorithms, can be adapted to use many parallel simulator instances. Furthermore, the authors were able to train using batch sizes considerably larger than the standard, without negatively affecting sample complexity or final performance. Within the results presented, Stoke and Abbel show that considered algorithms can indeed be adapted for running in parallel instances by applying minor changes. When looking at their experimental data across different tests (games, in their particular case), while performance remains acceptable in most cases, results are somewhat very different. This matches with the results observed by our team, which indicated that best configurations had to be found in order for these algorithms to obtain fast learning under similar performance levels.

\section{The robot\_gym framework}
\label{sec:robotgym}

\begin{figure}[ht!]
\centering
 \includegraphics[width=0.7\textwidth]{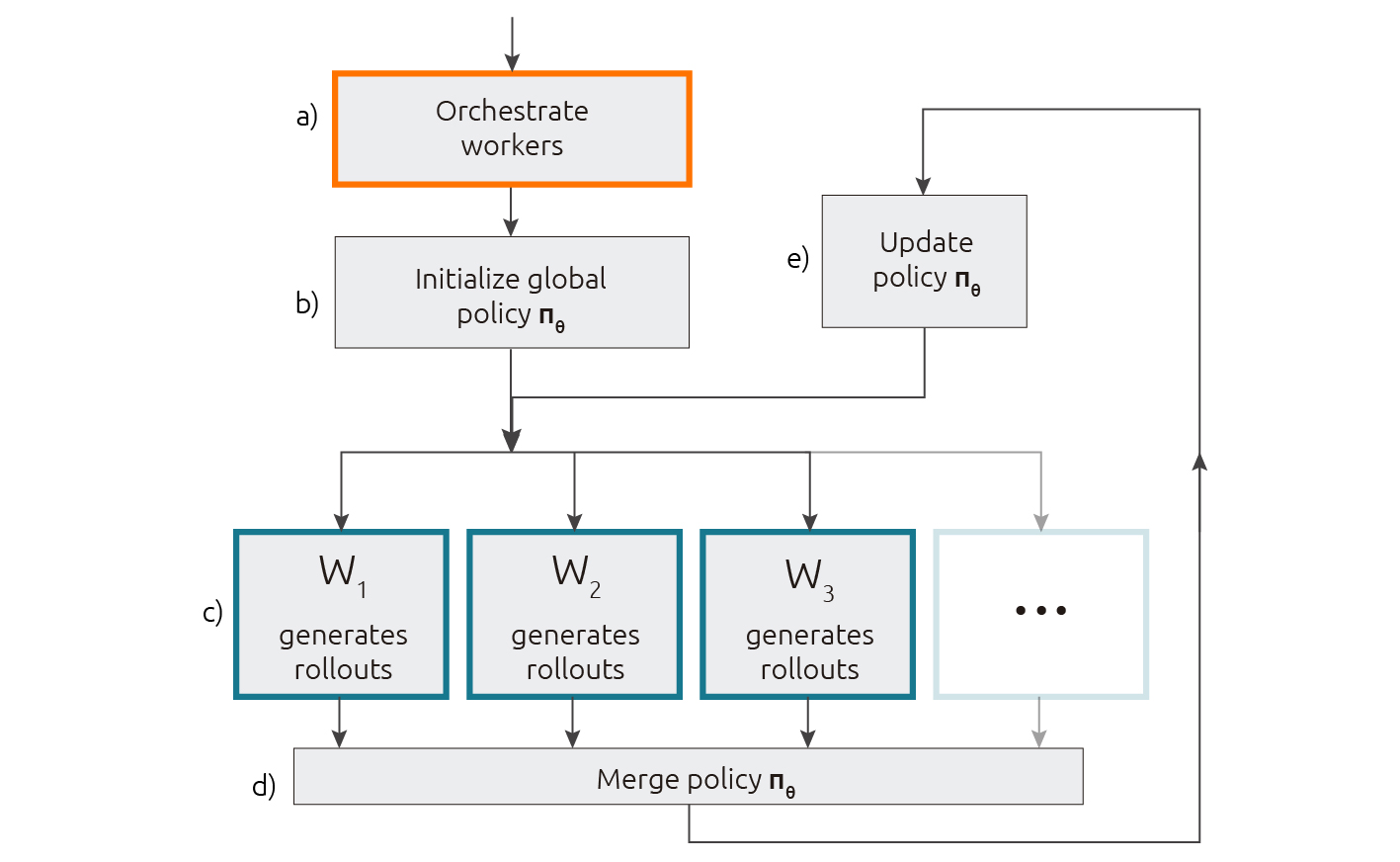}
\caption{\footnotesize The \emph{robot\_gym} architecture where a) pictures the worker orchestration, b) corresponds with the polity initialization c) represents each one of the workers, d) the policy merge task and e) the policy update process.}
\label{fig:robotgym}
\end{figure}

Most robots operate in a continuously changing environment which makes generalization of given tasks extremely hard. RL and, particularly, policy gradient methods, are among the techniques that allow for the development of more adaptive behaviors. However, even the most simple tasks demand long periods of training time. This aspect becomes specially relevant in robotics, where the time spent gathering experience from the environment (\emph{rollouts}) is significantly more relevant than the time spent computing gradients or updating the corresponding policy that is being trained.

To reduce the overall training time, this work proposes \emph{robot\_gym}, a framework for deploying robotic experiments in distributed \emph{workers}, that aims to reduce the time spent gathering experience from the environment and, overall, reduce the training time of robots when using RL. Figure \ref{fig:robotgym} pictures the architecture of the framework which has been inspired by previous work\cite{2017arXiv171209381L}. The following subsections describe the details of each block within the framework.



\subsection*{a) Worker orchestration}

\begin{figure}[ht!]
\centering
 \includegraphics[width=0.7\textwidth]{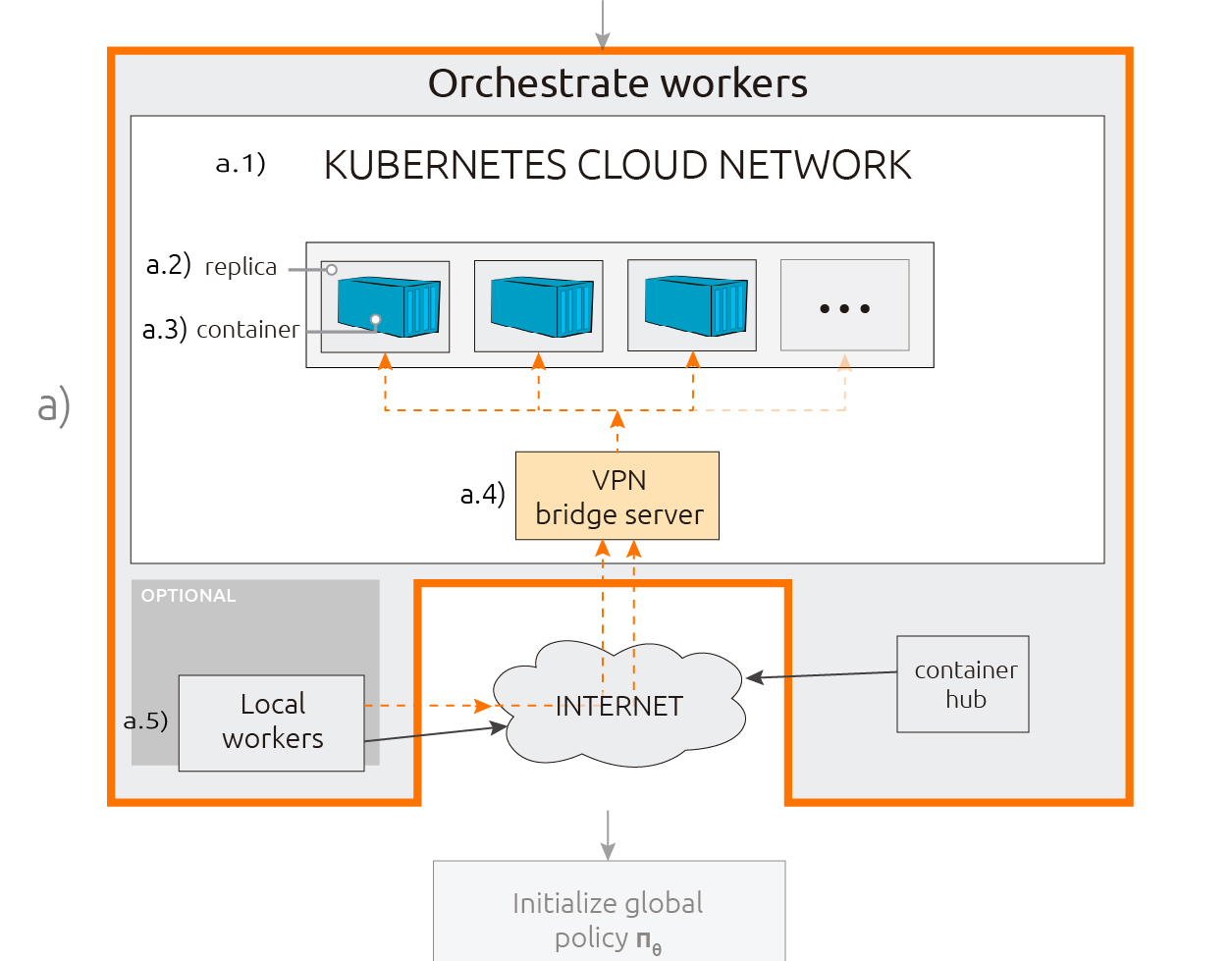}
\caption{\footnotesize Representation of the orchestration block from Figure \ref{fig:robotgym}. The figure describes the structure of the cloud network where a.1) represents the Kubernetes cloud network composed by a.2) replicas which contain a.3) containers. The VPN server a.4) coordinates and acts as the entry point for the cloud network. An optional local worker in a local replica is also presented in a.5).}
\label{fig:orchestration}
\end{figure}

When launching \emph{robot\_gym}, the orchestration block takes care of executing the different simulators that will perform rollout acquisition. Each individual simulator instance is typically known as a \emph{worker}. The number of resources required for a complete simulation of the robot allows us to allocate up to 4 workers per machine. Machines can be local or remote. To simplify the task of orchestrating workers, each one of them is containerized. To facilitate the deployment, we make use of Kubernetes\cite{brewer2015kubernetes}, an open-source system that allows to deploy, scale and manage container applications. Figure \ref{fig:orchestration} displays an architectural diagram of the orchestration process of \emph{robot\_gym}. An exemplary configuration file of Kubernetes is presented in the following list, where:
\begin{itemize}
    \item \texttt{replicas} represent the number of machines that will be launched among the available ones (local or remote). 
    \item \texttt{image} is the container that will be launched by each worker. 
    \item \texttt{resources} allow to define the computation limitations per machine.
\end{itemize}

\begin{code}
\caption{Kubernetes orchestration exemplary file.}
\begin{verbatim}
spec:
  replicas: 12
  ...
    spec:
      containers:
        - name: <container-name-preffix>
          image: <location-of-the-containerized-image>
          securityContext:
            privileged: true
          resources:
            limits:
              cpu: "4"
              memory: "3.8Gi"
            requests:
              cpu: "2.5"
              memory: "1.0Gi"
    ...
\end{verbatim}
\end{code}

Following from the Kubernetes configuration file, the Kubernetes cluster is created using a \emph{cloud service provider}. The provider will allow us to scale quickly the number of \texttt{replicas} composing a \emph{cloud network} (Figure \ref{fig:orchestration} \texttt{a.1)}). 

Each of the \emph{replicas} (Figure \ref{fig:orchestration} \texttt{a.2)}) composing the \emph{cloud network} will host one or multiple \emph{worker} containers. The number of containers that can run depends on the resources available at the \emph{replica} and the minimal resources requested per each container. These containers will (Figure \ref{fig:orchestration} \texttt{a.3)}) be fetched from a common place. All the instances need to run exactly the same code and version, each instance will fetch the latest available container image from a common container hub. A container hub is a service where the container engines fetch the containers that they should run. These hubs can be public or private, in this particular implementation, we use a private hub which only allows authorized clients to fetch the images.


The orchestration block should connect together all of the replicas and servers involved in the process, regardless of whether they are local or remote. To achieve this, a new replica dedicated to manage communications is created. This replica is not managed by Kubernetes in the same way the other replicas are. Instead, it will be running a Virtual Private Network (VPN)\cite{rosen2006bgp} server. Such server (Figure \ref{fig:orchestration} \texttt{a.4)}) will act as a communication bridge between local (Figure \ref{fig:orchestration} \texttt{a.5)}) and external replicas accessible from anywhere in a encrypted, protected and secure way. In addition, the VPN server allows to add new workers (remote or local) seamlessly to the \emph{cluster network}.



 
\subsection*{b) Policy $\pi_\theta$ initialization}

Following the orchestration, the robot \emph{robot\_gym} framework initializes a global policy $\pi_\theta$ that will be used by all workers while iteratively updated through the process as described in Figure \ref{fig:robotgym}.


By using Ray, a high-performance distributed execution engine, we are able to modify intrinsic parameters of the selected policy gradient technique. For example, we can define the batch size for the policy evaluations of different rollouts through \textit{rollout\_batchsize}, or via the number of Stochastic Gradient Descent (SGD) iterations on each outer loop by modifying \textit{num\_sgd\_iter}. Code listing \ref{listing:init} provides insight about the configuration options available for the policy $\pi_\theta$ initialization step.

\begin{code}
\caption{Policy gradient algorithm configuration file.}
\begin{verbatim}
DEFAULT_CONFIG = {
    # Discount factor of the MDP
    "gamma": 0.995,
    # Number of steps after which the rollout gets cut
    "horizon": 2000,
...
    # Initial coefficient for KL divergence
    "kl_coeff": 0.2,
    # Number of SGD iterations in each outer loop
    "num_sgd_iter": 30,
    # Stepsize of SGD
    "sgd_stepsize": 5e-5,
...
    # Batch size for policy evaluations for rollouts
    "rollout_batchsize": 1,
    # Total SGD batch size across all devices for SGD
    "sgd_batchsize": 128,
    # Coefficient of the value function loss
    "vf_loss_coeff": 1.0,
...              
\end{verbatim}
\label{listing:init}
\end{code}



\subsection*{c) Workers}

\begin{figure}[ht!]
\centering
 \includegraphics[width=0.8\textwidth]{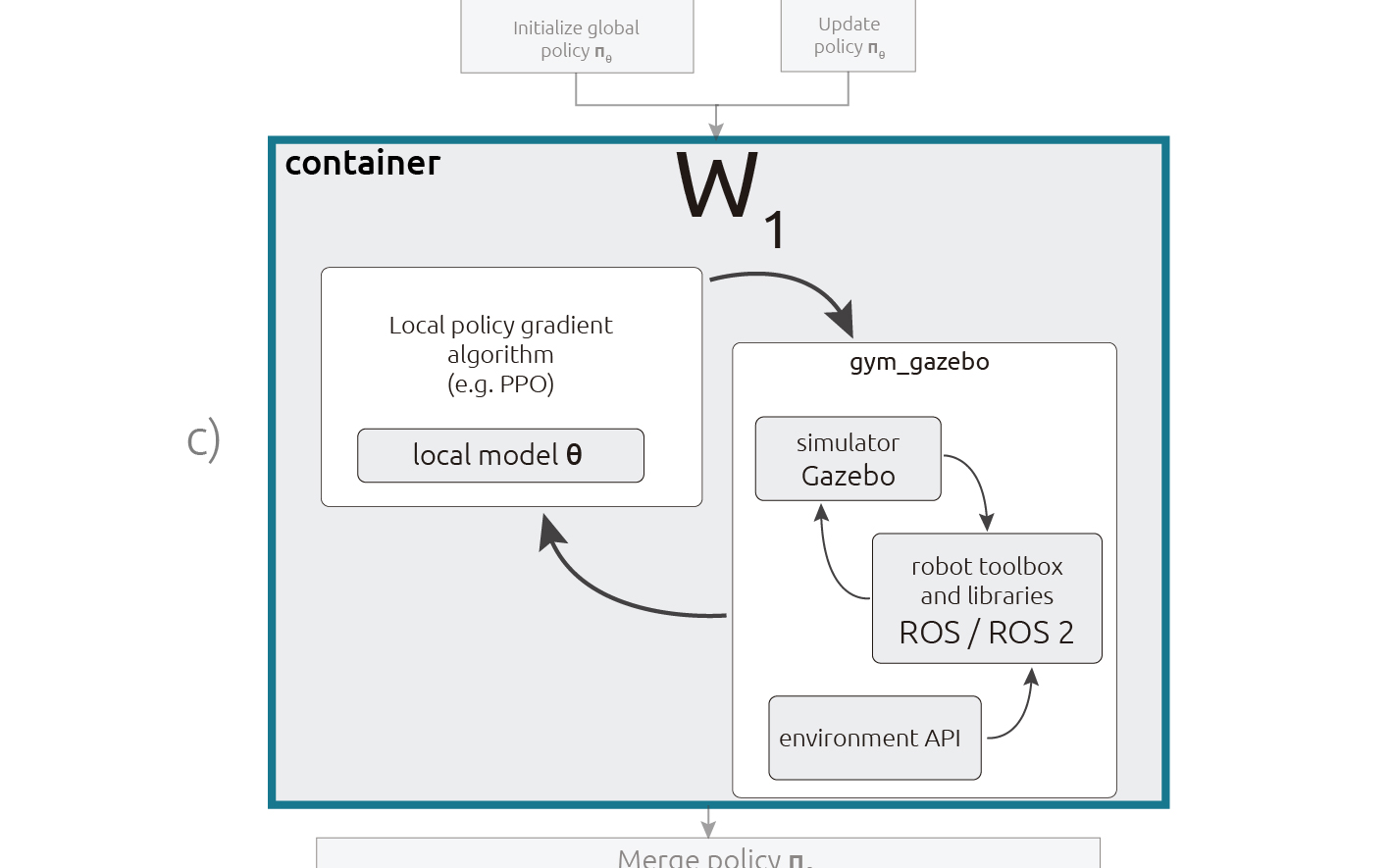}
\caption{\footnotesize Pictures a worker of the \emph{robot\_gym} framework.}
\label{fig:worker}
\end{figure}

Workers are containers that embed a local policy gradient algorithm and the \textit{gym\_gazebo} toolkit as displayed in Figure \ref{fig:worker}. \textit{gym\_gazebo} abstracts the robot environment providing a simple API that can be used by external algorithms. Within \textit{gym\_gazebo}, Gazebo and ROS\footnote{Latest prototypes include ROS 2 support.} are actively used. Gazebo provides a robust physics engine, high-quality graphics, and convenient programmatic and graphical interface; while ROS, a common toolbox that most roboticists are using nowadays, helps wiring everything together. Altogether, the architecture proposed allows roboticists to set up a virtual representation (simulation) of the robot through ROS packages that resemble or even is identical to the one used in the real robot. This facilitates the overall transfer of knowledge to the real robot, and empowers quick adaptations and changes in the robot configuration, critical when dealing with hardware.

Each worker collects \emph{rollouts} and improves its local model $\theta_{local}$. Such local model is then merged into the global policy as described by Figure \ref{fig:robotgym}.



%

\subsection*{d), e) Merge and update policy $\pi_\theta$}

With each worker producing their own \emph{rollouts} and updating their own local models ($\theta_{local}$ for each), the \textit{robot\_gym} framework, as pictured in Figure \ref{fig:robotgym}, needs to merge (\textit{d)}) and update (\textit{e)}) the local models $\theta_{local}$ into a global model $\theta$ to obtain a global policy $\pi_\theta$. The local model from each individual worker is pulled synchronously from remote workers; and then, merged together. This merging process is described in \cite{2017arXiv171209381L} and \cite{2017arXiv171205889M}\footnote{The merging process can be better understood by looking at Ray RL source code available at \url{https://github.com/ray-project/ray/tree/master/python/ray/rllib}.}. The updated model $\theta$ weights are then broadcast to all workers triggering a new cycle of the training process.



\section{Experimental results}
\label{sec:experimental}
To validate our framework, we ran experimental tests in simulation and deployed the results both in simulated and on real robots, obtaining similar results. The robots used in our experiments have been built using the H-ROS\cite{8046383} technology, which simplifies the process of building, configuring and re-configuring robots.

\begin{figure}[ht!]
    \centering
    \includegraphics[width=0.4\textwidth]{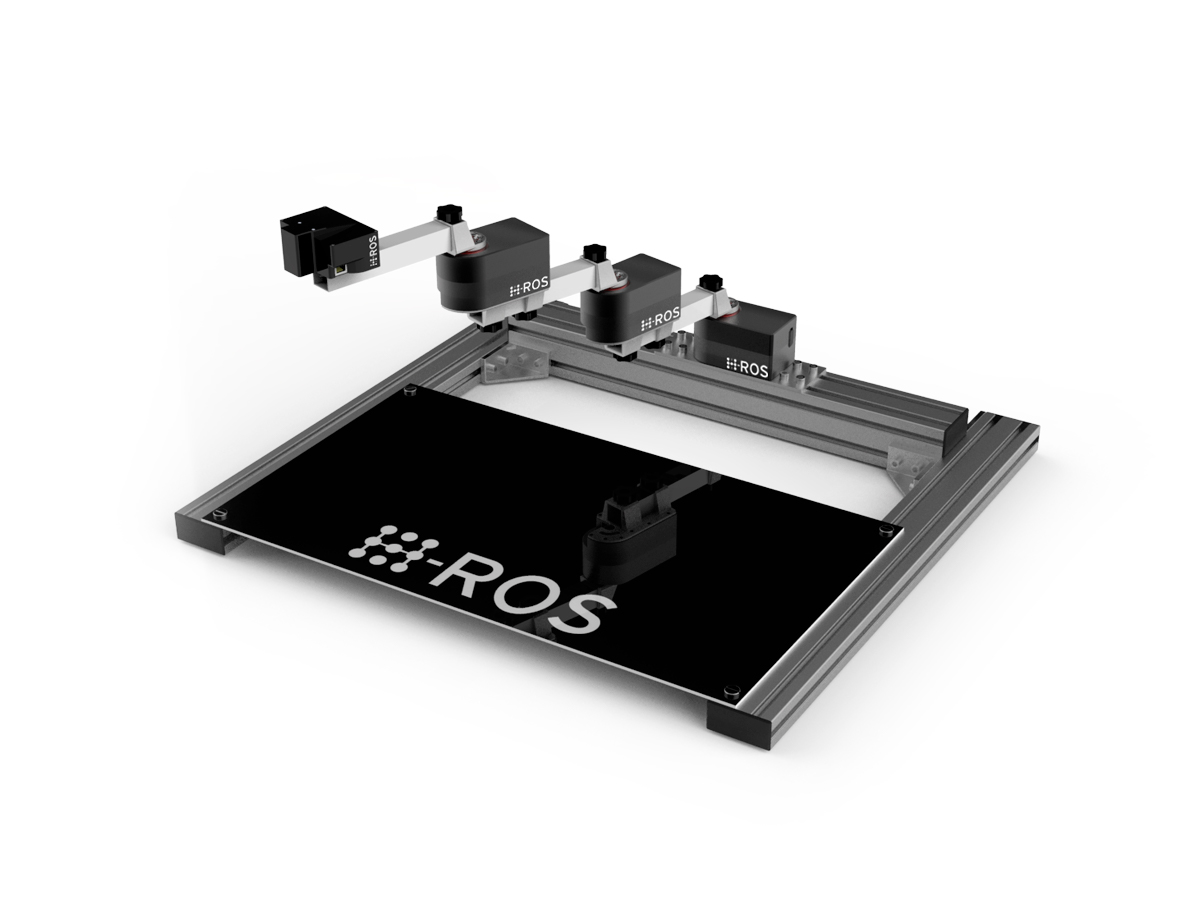}
    \qquad
    \includegraphics[width=0.4\textwidth]{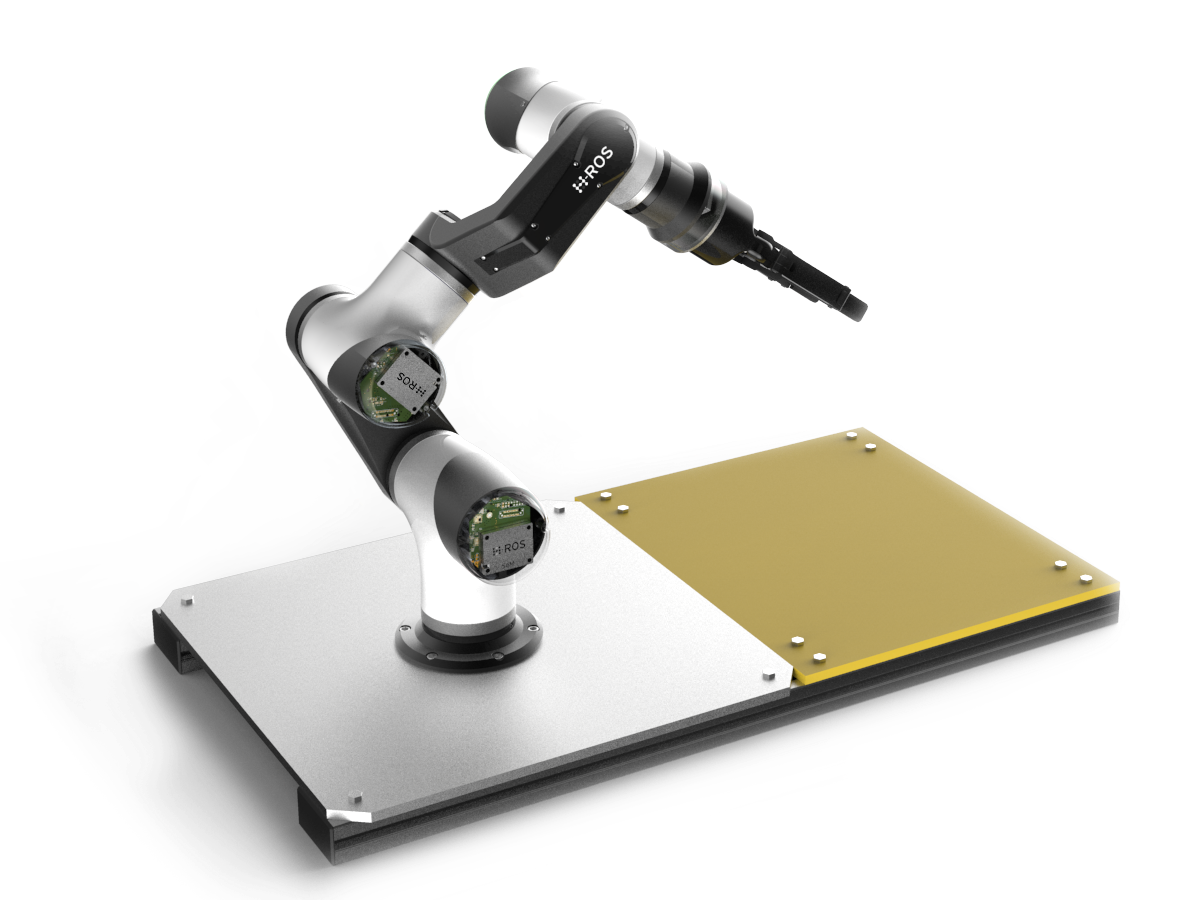}
    \caption{\footnotesize Modular robots used for the experimental testing. Left, a modular robot arm in a SCARA configuration with 3DoF. Right, a modular articulated arm with 6DoF.}%
    \label{fig:robot_arms}
\end{figure}

We experiment with two modular robots: a 3 Degrees-of-Freedom (DoF) robot in a SCARA configuration (Figure \ref{fig:robot_arms} \emph{left}) and a 6 DoF modular articulated arm (Figure \ref{fig:robot_arms} \emph{right}). We analyze the impact of  different number of workers, distributed through several machines and the number of iterations that the robot needs to converge. Our goal is to reach a specific position in the space stopping the training when the algorithm obtains zero as the target reward. Rewards are heuristically determined using the distance to the target point.

During our experimentation, we use the Proximal Policy Optimization (PPO)\cite{schulman2017proximal}, a state-of-the-art policy gradient technique which alternates between sampling data through interaction with the environment and optimizing the "surrogate" objective by clipping the policy probability ratio. 

\subsection{Modular robotic arm in a SCARA configuration (3DoF)}

We launched our experiment with 1, 2, 4 and 8 workers using 12 replicas. Within \emph{robot\_gym}, the use of ray library is in charge of distributing the workers between the available replicas. Some of the hyperparameters used during our experimentation are available in the code listing \ref{code:scara_params}.

\begin{code}
\caption{RAY parameters to train 3 DoF SCARA robot}
\begin{verbatim}
    ...
    # Number of steps after which the rollout gets cut
    "horizon": 2048,
    # Number of timesteps collected in each outer loop
    "timesteps_per_batch": 16000,
    # Each task performs rollouts until at least this
    # number of steps is obtained
    "min_steps_per_task": 2048,
    'num_sgd_iter': grid_search([50]),
    'num_workers': grid_search([1, 2, 4, 8]),
    'sgd_batchsize': 2048,
    ...
\end{verbatim}
\label{code:scara_params}
\end{code}

\begin{figure}[ht!]
    \centering
    \includegraphics[width=6cm]{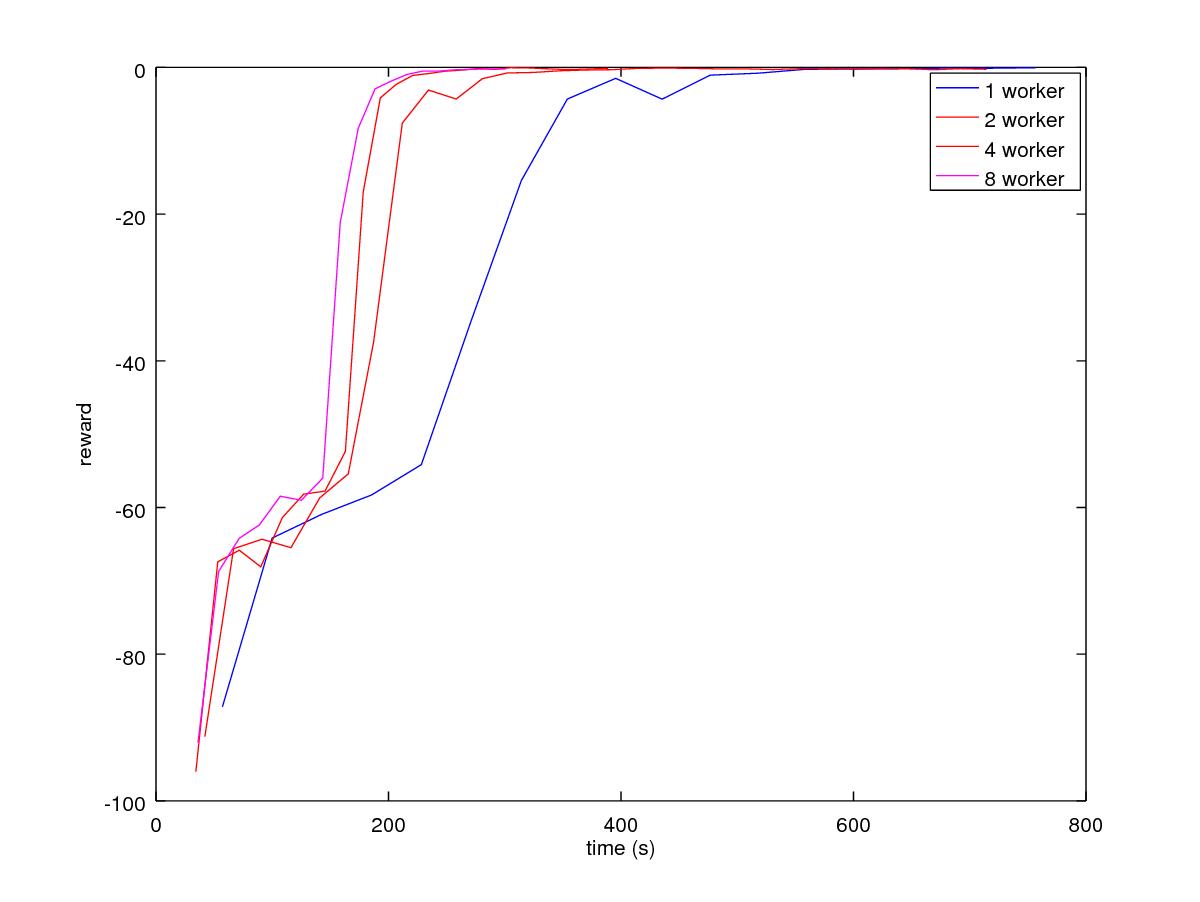}
    \qquad
    \includegraphics[width=6cm]{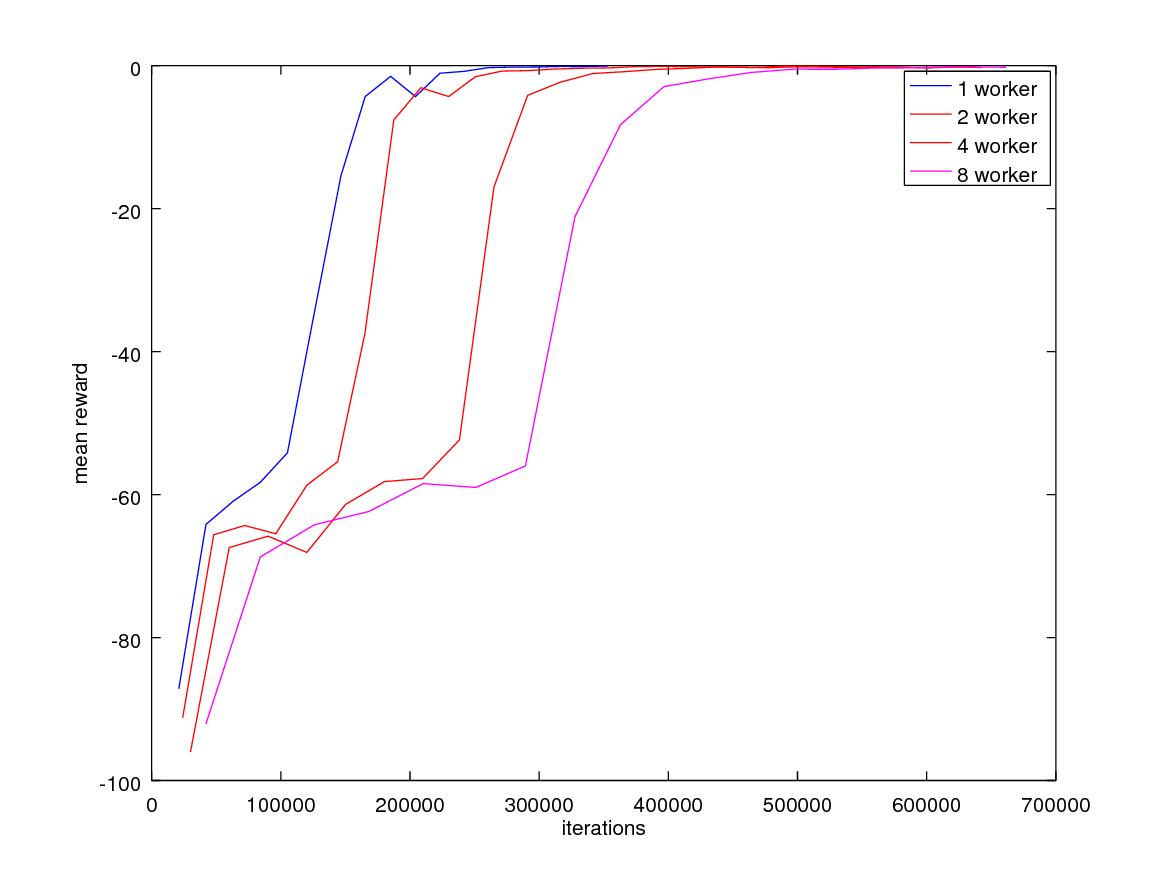}
    \caption{\footnotesize \emph{left}: "time" vs "reward" during approximately 700.000 iterations with different numbers of workers. \emph{right}: "iterations" vs "mean reward" with different numbers of workers.}%
    \label{fig:scara_learn}
\end{figure}

Figure \ref{fig:scara_learn} provides two representations: on the \emph{left} side, the reward obtained during training is presented and pictured against the time it took. Each curve illustrates the development of the same robot (modular robotic arm in a SCARA configuration with 3DoF) trained with the \emph{robot\_gym} framework, under a different number of workers distributed among the 12 available replicas. Using 1 worker, we can observe that the robot takes approximately 600 seconds to reach mean target reward of zero, that is, about 10 minutes of training time. Using 2 workers, the training time required to reach the target reward lowers to 400 seconds (6,6 minutes) approximately. When using 4, 8 or more workers, the training time required to reach a 0 target reward lowers to 300 seconds (5 minutes) approximately. From these results, we conclude that through the use of \emph{robot\_gym} framework and by distributing the rollout acquisition to a number of workers, we are able to reduce the training time by half in a simple robotics motion task.

The second graph (\ref{fig:scara_learn} \emph{right}) illustrates the mean reward obtained versus the number of iterations required. With this plot, we aim to visualize the sample efficiency of using a variable number of workers. As it can be seen, the more workers we use, the less sample efficient it becomes. This remains an open problem we have observed through a variety of different experimental tests comprising different simple robotics tasks.

We deployed the resulting global model $\theta$ into a) a simulated robot and b) a real robot. In both cases, the behavior displayed follows what was expected. The robots proceed to move their end effector towards the target point. Accuracy is calculated as the mean squared error between the target and the final end-effector when executing a trained RL network. The repeatability is calculated as mean square error between the mean from 10 experimental runs and the final end-effector position when executing a network.





\begin{table}[ht!]
\centering
\caption{\footnotesize Accuracy and repeatability (in mm) obtained in a 3DoF modular robot in a SCARA configuration when trained with the \emph{robot\_gym} framework for a simple robot motion task (reach a given point in the workspace).\\}
\begin{tabular}{l|c|c|c|c}
 \multirow{2}{4em}{\textbf{robot}} & \multicolumn{2}{|c|}{\textbf{accuracy}} & \multicolumn{2}{|c}{\textbf{repeatability}} \\
 \cline{2-5}
 & 1 worker & 8 workers& 1 worker & 8 workers \\
 \hline
 simulated & 2.80 & 1.37 & 4.52 & 3.57 \\
 real & 26.14 & 12.89  & 28.96 & 10.83 \\
\end{tabular}
\label{table:3dof}
\end{table}






\subsection{Modular robotic articulated arm 6 DoF}

In this second experiment, we train in simulation a 6 DoF modular articulated arm, as shown in Figure \ref{fig:robot_arms} \textit{right}. The objective remains similar: reach a given  position in space. However, in this case, the robot includes additional degrees of freedom, making  the search space much bigger; hence, the overall task more complicated, requiring additional training time. 

We launched our experiment with 1, 2, 4, 8, and 16 workers, using 12 replicas. From our experiments, we noted that this second scenario is much more sensitive to hyperparameters than the previous robot. Fine tuning was required for each different combination of workers, in order to make it converge towards the goal.


\begin{figure}[ht!]
    \centering
    \includegraphics[width=6cm]{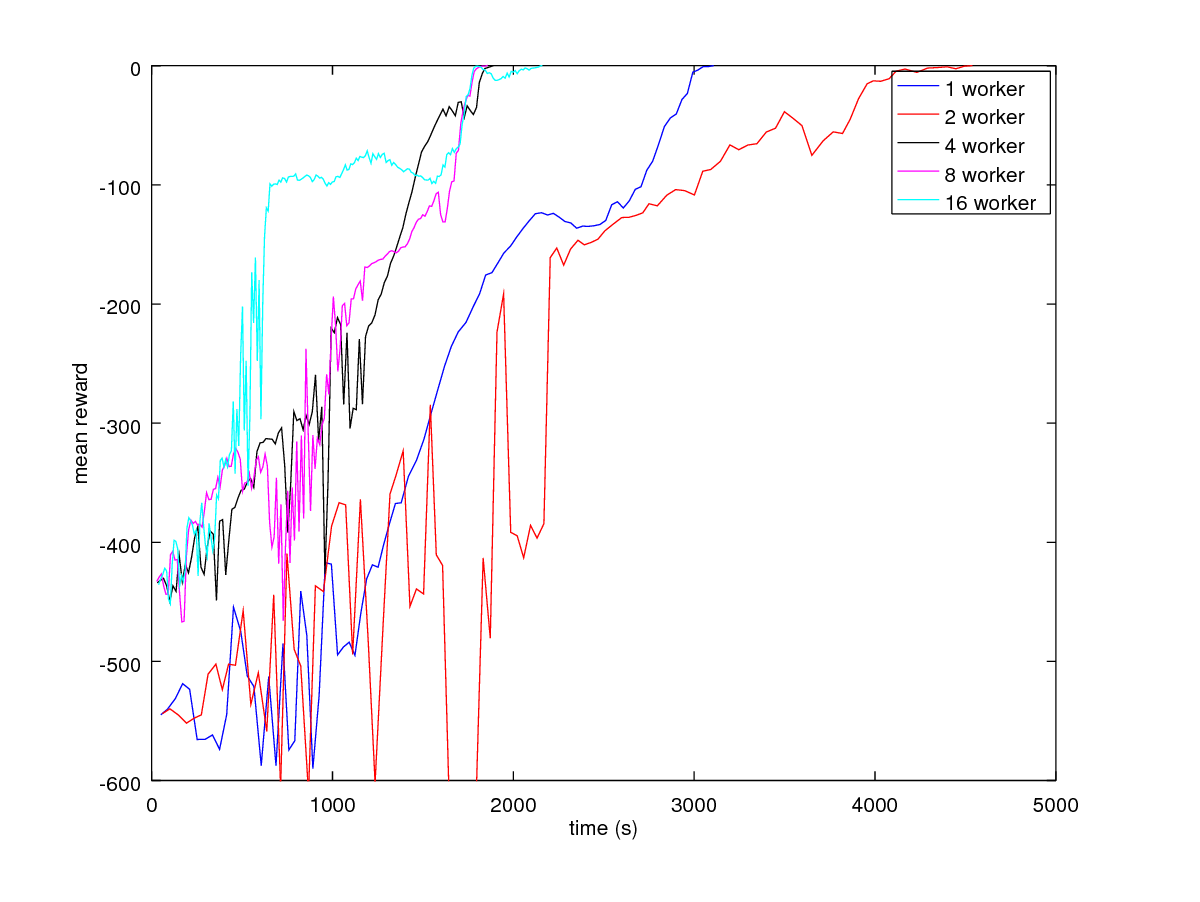}
    \qquad
    \includegraphics[width=6cm]{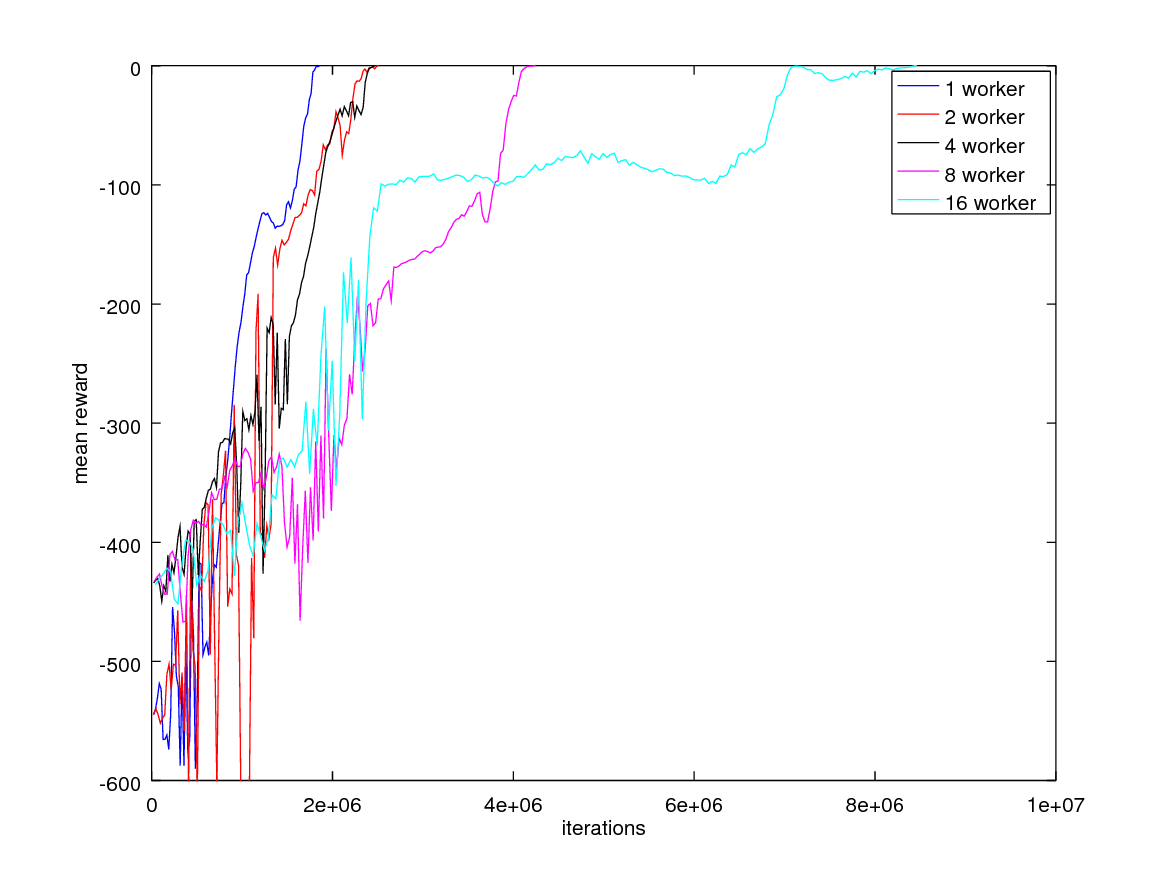}
    \caption{\footnotesize \emph{left:} 'time' vs 'mean reward' with different numbers of workers. \emph{right}: "iterations" vs "mean reward" with different numbers of workers for the 6DoF robot.}
    \label{fig:maira_learn}
\end{figure}

Figure \ref{fig:maira_learn} displays the results of training the 6DoF robot with the \emph{robot\_gym} framework. For a single worker, the time required to train the model until the mean reward reaches zero is about 3000 seconds (50 minutes). Adding additional workers (4 and 8), reduces the time required to reach the target down to 2000 seconds (33.3 minutes), or reduces the training time by more than 33\% compared to training only with a single worker.

\begin{table}[ht!]
\centering
\caption{\footnotesize Accuracy and repeatability obtained in a 6DoF modular robotic articulated arm when trained with the \emph{robot\_gym} framework for a simple robot motion task (reach a given point in the workspace).\\}
\begin{tabular}{l|c|c|c|c}
 \multirow{2}{4em}{\textbf{robot}}& \multicolumn{2}{|c|}{\textbf{accuracy}} & \multicolumn{2}{|c}{\textbf{repeatability}} \\
\cline{2-5}
 & 1 worker & 8 workers& 1 worker & 8 workers \\
 \hline
 simulated & 0.05 & 0.075 & 0.02 & 0.016 \\
\end{tabular}
\label{table:6dof}
\end{table}

As pointed out above, the hyperparameter sensitivity proved to be a challenge. The cases with 2 and 16 workers exemplify this fact. In the case of 2 workers, the time required to reach the training goal was the biggest, while the plots of the 16 workers' trials make us believe that additional parameter tuning may favor convergence towards the goal, further reducing the time required.

Similar to the previous experiment, the lack of sample efficiency remains an open problem to us. 




\section{Conclusions and future work}
\label{sec:conclusions}

This paper introduced \emph{robot\_gym}, a framework to accelerate robot training using Gazebo and ROS in the cloud. Our work tackled the problem of \emph{how to accelerate the training time of robots using reinforcement learning techniques} by distributing the load between several replicas in the cloud. We describe our method and display the impact of using different numbers of workers across replicas.


We demonstrated with two different robots how the training time can be reduced by more than 33\% in the worst case, while maintaining similar levels of accuracy. To the question \emph{how much does it cost to train a robot in the cloud?}, our cloud solution provider charged us a total of \EUR{214,80} for two weeks of experimentation work\footnote{This cost includes VAT and other costs deviated from the use of the services.}. 12 replicas were run on demand in parallel (only active if needed). 1603 hours of cloud computing was used in total (\EUR{0,134}/hour per instance or \EUR{1,606}/hour for all the replicas running at the same time).

The sample efficiency remains an open problem to tackle in future work.








\bibliographystyle{plainnat}
\bibliography{bibliography}

\end{document}